# Machine Learning Approaches to Real Estate Market Prediction Problem: A Case Study

Shashi Bhushan Jha[1], Vijay Pandey[2], Rajesh Kumar Jha[3], Radu F. Babiceanu[1, *]

[1]Department of Electrical Engineering and Computer Science, Embry-Riddle Aeronautical University, Daytona Beach, FL 32114, USA
[2]Department of Computer Science Engineering, IIT Kharagpur, India
[3]Department of Electronics and Communication Engineering, BNMIT, India

E-mail ID: jhas1@my.erau.edu, vijayiitkgp13@gmail.com, rajeshjnv23@gmail.com, babicear@erau.edu

*Corresponding author (phone: +1-386-226-7535; email: babicear@erau.edu)

**Abstract**

Home sale prices are formed given the transaction actors economic interests, which include government, real estate dealers, and the general public who buy or sell properties. Generating an accurate property price prediction model is a major challenge for the real estate market. This work develops a property price classification model using a ten-year actual dataset, from January 2010 to November 2019. The real estate dataset is publicly available and was retrieved from Florida's Volusia County Property Appraiser website. In addition, socio-economic factors such as Gross Domestic Product, Consumer Price Index, Producer Price Index, House Price Index, and Effective Federal Funds Rate are collected and used in the prediction model. To solve this case study problem, several powerful machine learning algorithms, namely, Logistic Regression, Random Forest, Voting Classifier, and XGBoost, are employed. They are integrated with target encoding to develop an accurate property sale price prediction model with the aim to predict whether the closing sale price is greater than or less than the listing sale price. To assess the performance of the models, the accuracy, precision, recall, classification F1 score, and error rate of the models are determined. Among the four studied machine learning algorithms, XGBoost delivers superior results and robustness of the model compared to other models. The developed model can facilitate real estate investors, mortgage lenders and financial institutions to make better informed decisions.

## 1. Introduction

Accurate home sale price prediction problem is essential for the real estate market. Besides buyers and sellers, the housing market includes several other stakeholders, local and state government, real estate dealers, financial institutions, and market predictors. Real estate domain is also an important part of the economy that can drive up and down the stock exchange market and even generate disruptive economic events. This was experienced not long ago, when the subprime mortgage crisis ultimately led to the depreciation of the real estate market and caused a worldwide recession [1]. The recession emphasized the need for better market prediction, particularly in the real estate domain. In periods of economic expansion, construction and employment in the real estate sector grow significantly and result in higher property prices [2]. The trend reverses in the periods of economic contraction. Nevertheless, accurate prediction of the real estate market would be significant for socio-economic development at local and national levels [3].

Machine learning (ML) was used in many domains for prediction purposes and should be an asset for real estate market prediction, as well. Properly trained algorithms could help with accurate home sale trends if market characteristics are accurately accounted for. This work addresses the home sale price with the aim to predict whether the closing sale price is greater than or less than listing price. The study uses the actual real estate market data from January 2010 to November 2019 retrieved from Florida's Volusia County Property Appraiser. The dataset includes 94,530 records and it is qualified by 21 important different features. In addition, the socio-economic factors such as GDP (Gross Domestic Product), CPI (Consumer Price Index), PPI (Producer Price Index), HPI (House Price Index) and EFFR (Effective Federal Funds Rate) are also collected for the same time period for housing price problems to enrich the dataset. To solve this problem, various powerful ML algorithms (Logistic Regression, Random Forest, Voting Classifier, and XGBoost) are applied for processing the dataset considering mean encoding or target encoding. The developed home price prediction model can assist the real estate investors, agents, fund managers, policy makers and mortgage lenders to make better informed decisions.

The remaining sections of the paper are structured as follows. Section 2 reviews the literature on the home sale price prediction when employing ML techniques. The problem description and data analysis, including dataset feature analysis, is presented in Section 3. Section 4 presents the prediction mode design and implementation along with various ML algorithms trained to address the home sale price problem. The model empirical results are outlined and discussed in Section 5. The paper concludes with Section 6 where the main benefits of the study are emphasized, and distinct future research is proposed.

## 2. Literature Review

The literature covering sale price prediction approaches within different markets was surveyed and reported recently [4]. It was categorized in separate areas, approach that this work agrees with. Our study looks specifically to the real estate market and the reported ML approaches for sale price prediction and leaves out other types of prediction techniques.

### 2.1. Studies on ML approaches to real estate price problem

In a similar framework to our study, Park and Bae [1] addressed the real estate price classification problem by considering Virginia's Fairfax County as case study. But the study was limited to only one type of property, so the dataset was considerably lower than this current study. Also, the regional characteristics of the two counties are substantially different. which means the and having different regional characteristics. This means that the results can only be applied for the specificities of that study. Naive Bayesian, AdaBoost, and RIPPER approaches were used to develop a home price classification model. The reported future research directed the attention to the inclusion of appraised value of a properties and property tax, along with an increased dataset. In another study, a hybrid of genetic algorithms and support vector machines method was built to generate a home price prediction model [3]. Support-vector machines are supervised ML models that can be used as classifiers, so addressing the same type of problem as the current study. No information is reported on the size of the dataset, so no meaningful comparison can be made with our current study. Plakandaras et. al. [2] addressed the real estate property price index problem, however at a macroeconomic scale. The entire U.S. market is considered, employing 11 macroeconomic variables, and covering more than 100 years of aggregated price data, not actual home sale data. The model combines the Ensemble Empirical Mode Decomposition with Support Vector Regression and its prediction is compared with Random Walk, Bayesian Vector Autoregressive, and Bayesian Autoregressive models.

A case study analysis on the same regional area as our current study has the objective to predict the home price valuation using ML techniques based on regression modeling [4]. It uses a 5-year dataset of approximately 50,000 cured data points and employs XGBoost, CatBoost, Lasso, Voting Regressor, and other techniques to predict the home prices and compare them to a benchmark model. Due to the large dataset, the model accuracy increases by 10 percent compared to other reported ML approaches. Another study, on a 11-year dataset with an initial 50,000 data points was conducted with selected features such as usable area, building year, number of rooms, number of stories in the building, and geographical coordinates from the city center and to the nearest shopping center [5]. The learning models considered included Support Vector Regression, Generalized Regression Neural Networks, Adaptive Neuro-Fuzzy Inference Systems, and Genetic Fuzzy Systems.

Truong et. al. [6] used a 10-year signficantly large dataset to validate multiple regression techniques and deliver solutions for the home sale price prediction problem. The initially more than 300,000 datapoints exhibiting 26 features were cured down to a little bit more than 230,000 datapoints and 19 features, among them location, area, population, and HPI. The datset was processed through five different regression approaches, XGBoost, LightGBM, Random Forest, Hybrid Regression, and Stacked Generalization, the last one by itself being a stacked ensemble of Random Forest and LightGBM, complemented by XGBoost on the second stacking level. Another relatively large study on more than 20,000 data points in a different regional area was conducted to assess the spatial dependency of real estate price appreciations, and certain influential variables such as house structural attributes, locational amenities, visitor patterns, street view images, and socioeconomic attributes of neighborhoods, and their relationships [7]. Through deep learning the model attempts at price appreciation potential by fusion multiple data sources. Lastly, the review of the literature identified a series of three multi-target support vector regression models via correlation regressor chains designed for analyzing 24 multi-target multi-domain datasets [8]. The largest of the datasets is a 20,640 regional housing data points with seven features and two targets. The results of the multi-target model are statistically analyzed and compared with seven state-of-the-art methods.

**2.2 Research gap and contributions**

Over the past decade, most of the research on the real estate market problem addressed home price problem using hedonic-based regression models. ML techniques were employed in a relatively small percentage of the reviewed literature, and they are references above. One aspect inadequately addressed is the inclusion of socio-economic factors in the proposed models. Given the above findings, this work focuses on modeling the home sale price problem with ML algorithms and considering socio-economic factors such as GDP, CPI, PPI, HPI, and EFFR. Furthermore, with respect to methodology, the mean target encoding is used to enhance the performance of the prediction models.

## 3. Problem Description and Data Analysis

This section details the home sales prediction problem and includes the dataset information and data preparation process. Elements of feature engineering are presented along with the importance ranking of the feature analysis process. The section also includes the rationale for enlarging the feature suite besides the ones extracted from the dataset.

### 3.1 Dataset Collection

Volusia County, located in the east-central part of the state of Florida, is one of the well-developed counties in the state. The dataset of the real estate market is updated weekly by Florida Department of Revenue Substantive Regulations and is publicly available[1]. The collected dataset consists of 94,530 records covering almost ten years of transactions (January 2010 to November 2019) and which includes more than 100 variables. In the current study, after consulting with real estate domain experts, only 21 important features (variables) of the dataset are considered. Furthermore, since the dataset covers the early 2010s years after the 2008 recession, during the housing market revival time, this study also considers key socio-economic factors of the housing market, such as GDP, CPI, PPI, HPI, and EFFR. Therefore, the total number of features for the considered home price problem is 26. By including the sale price as well, the number of variables under study is 27.

The economic activity of a region is estimated by its GDP, which is the aggregate value of all services and final goods produced during a study period. The price index for consumer (CPI) is a metric that explores the weighted average prices of consumer services, such as medical care, food, and transportation, while the producer price index (PPI) is an estimate of mean price provided by producers of domestic services. HPI, the index measuring the property price estimates the average changes in property prices based on several factors such as sales, mortgages, etc. The HPI dataset is publicly available for research at the national, state, and county levels. The last of the five metrics noted above, EFFR, is derived from what is called Federal Funds Rate (FFR), an interest rate banks are using for lending reserve balances to the depository organizations. Then, the EFFR is the weighted average interest rate that borrowing banks pay to lending banks to borrow funds.

### 3.2 Data Preparation

The target variable in the problem, the sale price denoted as *price_high_low*, is generated with the help of two such features, the *aprtot*, defined as the appraised value of the property suggested by the Government, and *price*, which is the actual sale price of the property. The *price_high_low* variable is assigned the value 1 if the property's actual selling price is greater than the total appraised value of the property, and 0, otherwise. Table 1 presents the dataset description with the mean and standard deviation calculated for 25 of the variables.

---

[1] http://vcpa.vcgov.org/

Table 1. Description of variables

| Variable Name | Description | Mean | Standard Deviation |
|---|---|---:|---|
| *parid* | Parcel ID (property ID) | 4.35E+06 | 1.68E+06 |
| *aprland* | Total land just value | 4.23E+04 | 6.68E+04 |
| *aprbldg* | Total building(s) just value | 1.61E+05 | 8.89E+04 |
| *aprtot* | Just value at time of sale or total just value | 2.04E+05 | 1.34E+05 |
| *nbhd* | Neighborhood code | 3564.314015 | 1585.211332 |
| *rmbed* | Number of bedrooms | 2.952666 | 0.761134 |
| *sfla* | Square footage of living area | 1721.419501 | 678.18781 |
| *total_area* | Total building square footage | 2499.248978 | 972.172054 |
| *yrblt* | Year built | 1988 | 21 |
| *misc_area* | Miscellaneous area (gym, pool, parking) | 172.508538 | 341.062468 |
| *zip21* | Zip code of area | 32376 | 286 |
| *sale_date* | Sale date of property/parcel | - | - |
| *sasd* | School assessed value | 1.84E+05 | 1.27E+05 |
| *nsasd* | Non-school assessed value | 1.83E+05 | 1.27E+05 |
| *stxbl* | School taxable | 1.66E+05 | 1.29E+05 |
| *nstxbl* | Non-school taxable value | 1.51E+05 | 1.30E+05 |
| *cotxbl* | County taxable value | 149546.6 | 1.31E+05 |
| *citxbl* | Sale price of house city taxable value | 1.15E+05 | 1.23E+05 |
| *price_high_low* | Sale price of house | 0.491501 | 0.499931 |
| *gdp* | Gross domestic product (GDP) | 18753.09591 | 1842.06229 |
| *cpi* | Consumer price index (CPI) | 294.403424 | 24.815653 |
| *ppi* | Producer price index (PPI) | 197.366783 | 7.040979 |
| *hpi* | House pricing index (HPI) | 377.019444 | 71.960126 |
| *effr* | Effective federal funds rate (EFFR) | 0.818461 | 0.840287 |
| *hx_flag* | Flag of homestead properties | - | - |
| *luc* | Property class (PC) | 147.841026 | 402.37172 |
| *mararea* | Market area | 10.741929 | 5.814958 |

The main purpose of this study is to classify whether the property sale price is higher or lower than the appraised price. To solve the problem, this research employs a number of ML techniques discussed in the next sections.

### 3.3 Feature Analysis

Feature analysis was conducted on the dataset and exemplified here in several plots, particularly covering the sale price, *price_high_low*, and the following variables *gdp*, *cpi*, *hpi* and *effr* over the entire period of study, 2010 to 2019. In addition, the feature importance is described with the help of several other plots. All plots are generated using ML based techniques such as Random Forest and XGBoost Classifiers. The number of home sales over the entire study time is depicted in Fig. 1. It can be observed that home sales are increasing over the years, however, the sales in initial years were low due to the recession just before the start of the study period. Also, the chart shows a lower number of sales in 2019 since data collected covers only 11 months.

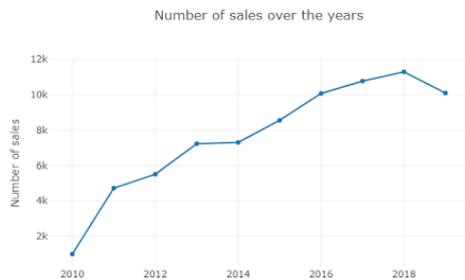
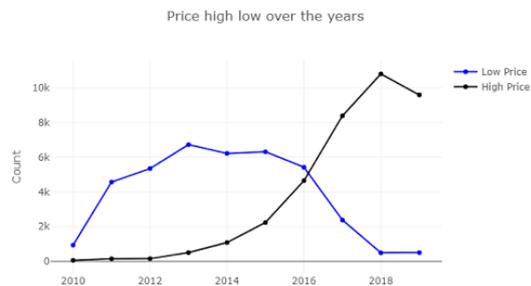

Fig. 1. Number of home sales over the study period

Fig. 2. Home sale price high and low

The retrieved dataset shows that for transactions from 2010 to 2016 the appraised price was higher than the actual sale price of the property. This behavior, shown in Fig. 2, was attributed to the change in the home market in the revival time after the recession. However, after 2016 there was a sudden increase in the number of higher sale prices. As the economy was getting stable, the activities in the real estate market boosted up and the real estate property market was gaining as investors and mortgage lenders showed increased interest. This observation provided the rationale to consider the economic factors selected for feature analysis (*gdp*, *cpi*, *hpi* and *effr*). Further study showed that these factors are not only highly correlated with the target variable, behavior shown in Fig. 3 and 4, but they enhance the performance of the model, which is presented later in the results section. In contrast, the variable associated in the model with the producer index, *ppi*, does not show correlation with the target variable, which is noticeable in the Pearson Correlation plot of Fig. 5. As a direct implication, *ppi* was not included along with the other variables in the plots. For visual illustration purposes, the values of each factor in Fig. 3 generated for the sale price, *price_high_low*, *gdp*, and *cpi* are scaled so that the comparison of the variable can be performed on the [0, 1] interval. It can be inferred that as the *gdp* and *cpi* value sharply increase, home sale price grows as well. The same approach was used for the values of the factors *price_high_low*, *gdp*, and *cpi*, plotted in Fig. 4. Their values are scaled so that the comparison can be performed in the [0, 1] range. It can be observed from Fig. 4 that sale price is highly correlated with the *hpi* and *effr* variables. As the hpi and effr values increase, the sale price also increases.

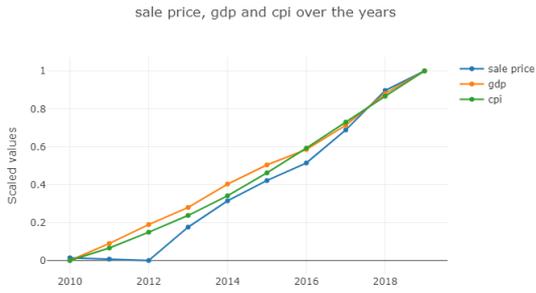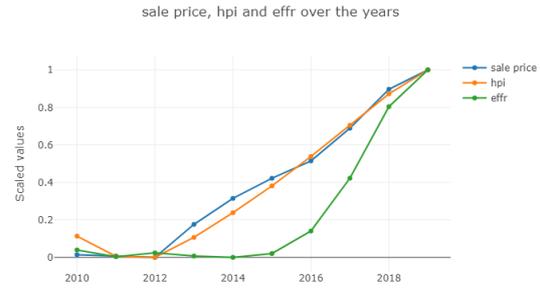

Fig. 3. Sale price, GDP, and CPI over the study period

Fig. 4. Sale price, HPI, and EFFR over the study period

The Pearson correlation plot [9] of Fig. 5 was generated to determine and present the correlation among the variables selected for the home price problem. Twenty-seven 27 variables are considered, including the target variable *price_high_low*, and their correlation are determined. However, in the proposed prediction model, this variable will be excluded because its value is unknown before running the model. The plot of Fig. 5 shows that the target variable is significantly correlated with *gdp*, *cpi*, *hpi* and *effr* (darker green color). Other variable show negative correlation (different shades of red), such as age in relation to year built, total area, and square foot of living area.

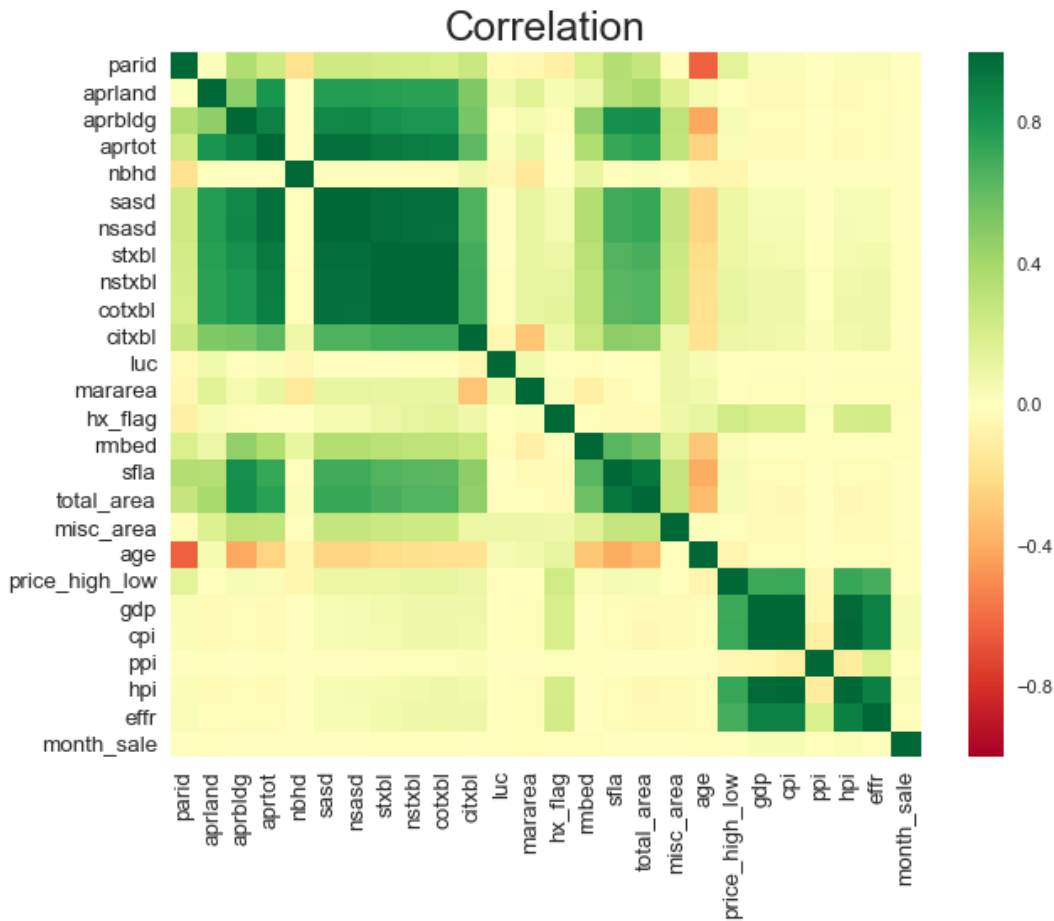

Fig. 5. Pearson correlation across the selected features

**3.4 Feature Importance**

Feature importance is analyzed using widely recognized ML techniques such as Random Forest Classifier and XGBoost Classifier and the analysis results are plotted in the next three charts. Feature importance techniques assign a value to each input based on their relative importance. The Random Forest Classifier model analyzes 24 out of the 27 variables, leaving out the *price_high_low*, *parid*, and *sale_date*. The output of the model is sorted in descending order according to feature usefulness. Fig. 6 highlights the important features identified by this classifier *cpi*, *effr*, *hpi*, *gdp*, *hx_flag* and *age*. Variables like the land use code (*luc*) result as having the least importance for this classifier model.

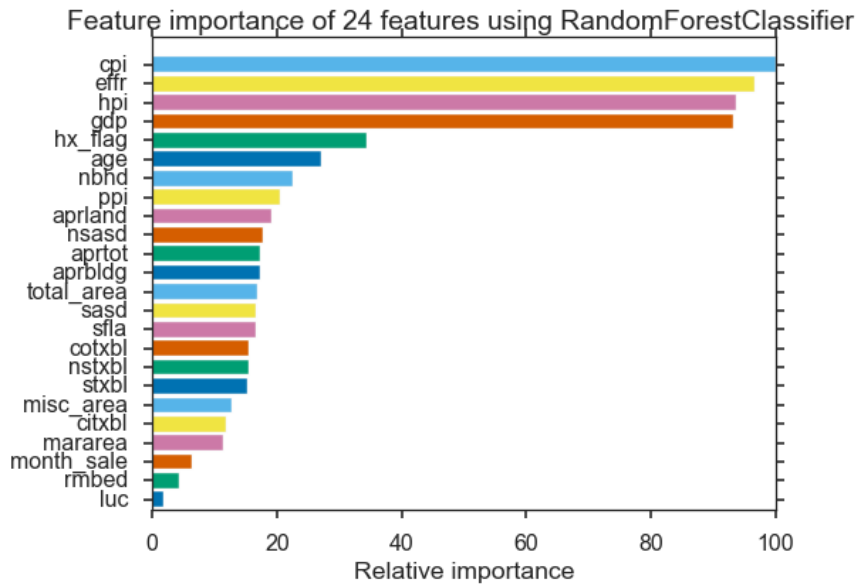

Fig. 6. Feature importance using Random Forest Classifier

Fig. 7 depicts the feature importance generated using the XGBoost classifier. It can be observed that *gdp* and *cpi* are the most influential factors in the XGBoost analysis, while *month_sale* is shown as having the least importance.

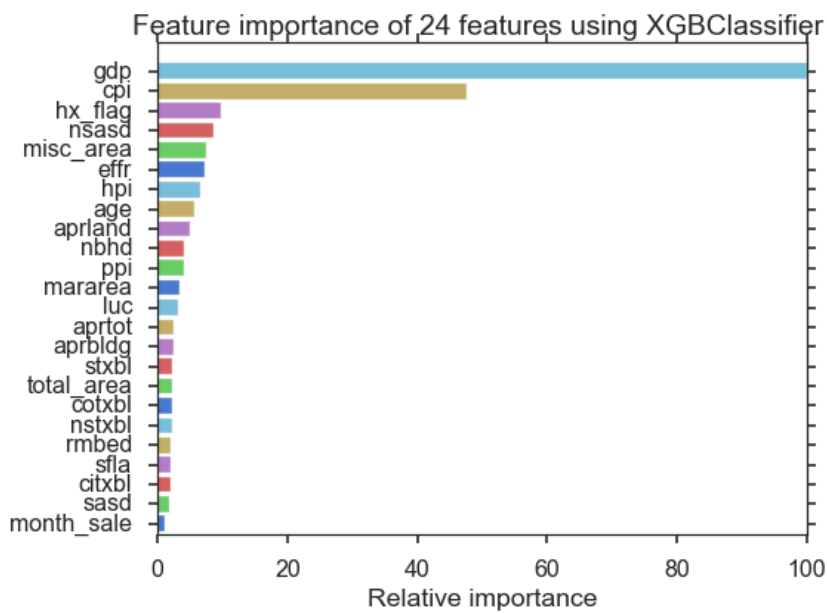

Fig. 7. Feature importance using XGBoost Classifier

To further analyze the dataset through feature engineering, the mean encoding process is employed for the Random Forest Classifier. The total number of features increases because mean encoding is performed on a set of features and results in the creation of new features *F*1 and *F*2, which are the predicted values of socio-economic factors and real estate market dataset, respectively. In addition, sales and the property build year also retain distinct features such as *ys*1, *ys*2, *yb*1 and *yb*2. More detailed explanation of the mean encoding is given in the methodology section. Fig. 8 shows that *F*2, *cpi*, *effr*, and *gdp* are ranked high in importance after employing mean encoding. However, *luc* (land use code), *rmbed* (number of bedrooms) have the least important features.

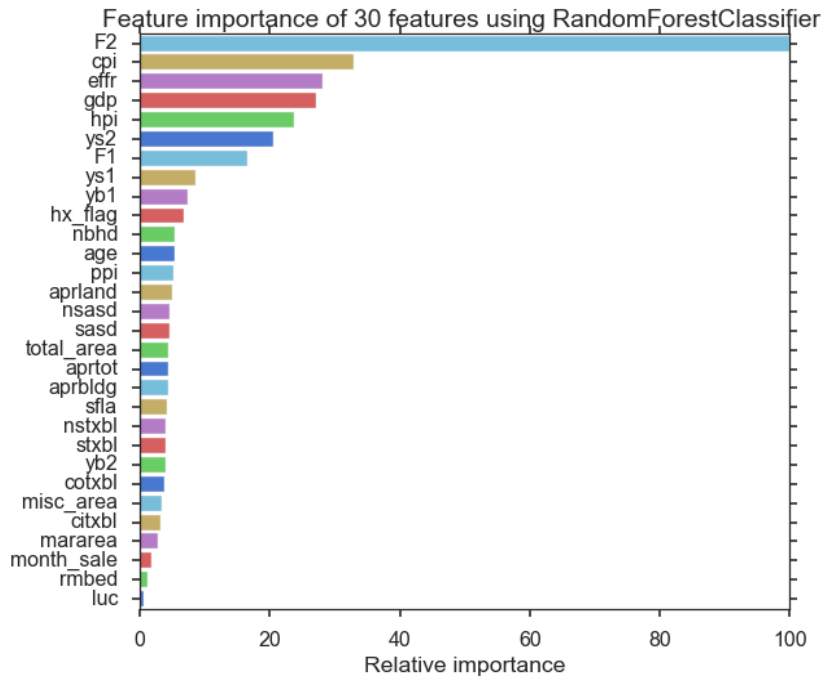

Fig. 8. Feature importance after mean encoding using Random Forest Classifier

**4. Prediction Model Design and Implementation**

The feature engineering section shows that a few independent variables are linearly correlated with the target variable while others are not. To capture the socio-economic factors for the home sale price problem, several variables are considered monthly rather than yearly. The monthly dataset is retrieved from a Federal Reserve data repository[2]. Incorporating socio-economic factors with the real estate market dataset provides an enhanced image of the society, which carries influence on real estate market transactions. This section surveys the ML algorithms employed on the enhanced dataset. Every ML algorithm has its own advantages and disadvantages, highly dependent on the dataset to be fed into the algorithms. Broadly, ML algorithms are divided into predefined classes, such as hyperplane-based, tree-based, gradient learning-based, bagging methods, boosting methods, linear algorithms, non-linear algorithms, Bayesian learning, and nearest neighbor-based algorithms. This work employs Logistic Regression, Random Forest, XGBoost and Voting Techniques to solve the home sale price problem. The important details of each of the four algorithms selected are discussed in detail in the next sections.

---

[2] https://fred.stlouisfed.org/

## 4.1 Algorithms for Prediction

*4.1.1 Random Forest (RDF)*

RDF [10] is a collection of decision trees where each tree is built from a sample drawn from the training set. The algorithm uses feature randomness when building single trees, such that it results in an uncorrelated forest of trees whose classification outcome is better than of any single tree. To increase randomness in the problem, a subset of given features are considered for splitting, with operations performed on each tree node [11]. During processing, individual decision trees could be characterized by high variance, which can lead to overfitting of the tree estimator. A solution to address the high variance is to provide two types of randomness. Random subset samples result in different errors, but those are cancelled out by predicting the class the having majority votes. While random forests may result in an increased bias, it is the variance that needs to be addressed first, rather than the bias.

*4.1.2 XGBoost*

XGBoost [12] stands for "Extreme Gradient Boosting" and is built on the concept of gradient boosting trees [13]. Objective function differentiates the algorithm from other gradient boosting techniques. It includes two parts, the training loss, and the regularization term, as seen in the Eq. (1) below.

$$L(\emptyset) = \sum_i l(\hat{y}_i, y_i) + \sum_k \Omega(f_k) \quad (1)$$

$$\text{where, } \Omega(f) = \gamma T + \frac{1}{2}\lambda \|w\|^2$$

The training loss measures the degree of prediction success with respect to the training data, while the regularization term controls the complexity of the model and helps with model generalization. The Taylor expansion of the loss function up to the second order is used to expand the loss function. In practice, XGBoost performs very well when applied to hardware and software optimization techniques. In addition, it comes with the significant advantage of performing faster computations with lower amount of computing resources needed.

*4.1.3 Voting Classifier*

Voting Classifier [14] combines different ML algorithm-based estimators and returns the class having majority votes among participating estimators. A voting classifier is a type of technique that considers base classifiers and fits them on the dataset. This type of classifiers is useful for a set of equally well performing estimators to balance out individual weaknesses. The method performs best when the predictors are as independent from each another as possible. However, it uses different algorithms to train each classifier, which increases the chance they will result in different types of errors.

*4.1.4 Logistic Regression Classifier*

Logistic Regression Classifier [15] is a linear model class based on the likelihood estimate of the true class. The logistic function is used to output the probability of each class, and cross entropy loss is used as a cost function. The algorithm is the variant of linear regression for classification problems. The main difference between the two is the use of logistic/sigmoid function to produce the output. The output of logistic regression lies between 0 and 1, whereas output of linear regression can be any real number. High dimensional sparse dataset having linearly separable classes are best suited for this algorithm to work.

**4.2 Prediction Model Performance Measures**

The performance metrics for classification models are based on the accuracy, F1 score, precision, recall, macro averaging, and micro averaging. The last two are used if there are more than two classes in the target variable, so the proposed prediction model disregards them. The accuracy metric is preferred if target classes are balanced, and since the used dataset has balanced classes, this is one of the performance measures considered. The model considers the other three metrics, precision, recall and F1-score to measure the performance in a more robust way. In addition, 10-fold cross validation technique is used to compare the extrapolation ability of the model, which will be tested on 10 disjoint validation sets and their average calculated to determine the final score.

For the feature selection method, this work considers the 5-fold cross validation on the training data to decide a relative optimal set of predictor variables. The process starts by selecting all variables for modeling, followed by running the feature importance method to obtain the important variables until the evaluation criteria of internal cross validation reaches maximum. Then, the selected subset of variables is considered in the outer evaluation. During the process of model construction, certain ML algorithms have hyper-parameters that need to be tuned, such as the tree depth, regularization parameter for XGBoost, number of trees and sample size for each split in random forest. For the parameters, the grid search method is further applied in the innermost 5-fold cross validation, and the parameters with optimal performance measure metrics value are selected to train the model. The specific technique employed to build these ML models and tune the meta-parameters consider the Scikit-learn package from Python platform [16] . While training and testing the model, and to show model performance, advanced visualization approaches are considered. For example, Yellowbrick, a Scikit-learn package [17] is used for training and testing the model. The detailed interpretation of each visualization graph is discussed next.

*4.2.1 Receiver Operating Characteristic (ROC) Curve*

ROC curve [15] is a graphical plot used to show the diagnostic ability of binary classifiers. It is plotted with False Positive Rate (FPR) against the True Positive Rate (TPR). The model considers different threshold values and finds the true positive and false positive classes keeping these values as the boundary for the two classes. The observed values are plotted as a ROC curve where true positive rate is on the *y*-axis and false positive rate is on the *x*-axis. ROC curve with respect to RDF and XGBoost algorithms are depicted in Fig. 10 and 15, respectively.

*4.2.2 Kolmogorov-Smirnov (K-S) Statistic Plot*

The K-S chart [18] measures performance of classification models by comparing the distribution of two classes and assessing if they are properly separable or not. Although the test is nonparametric, it does not assume any particular underlying distribution. More accurately,

K-S test is a measure of the degree of separation between the positive and negative distributions. The K-S value is 1 if it partitions the population into two separate groups in which one group contains all the positives and the other all the negatives. On the other hand, if the model cannot differentiate between positives and negatives, then it is as if the model selects cases randomly from the population, in which case the K-S value is 0. In most classification models, the K-S test falls between 0 and 1, and the higher the K-S value, the better the model is at separating the positive from negative cases. K-S chart of RDF and XGBoost algorithms are presented in Fig. 11 and 16, respectively.

*4.2.3 Cumulative Gain Curve*

Cumulative gain [19] is a visual aid for measuring model performance used to find the effectiveness of a binary classifier. It consists of a lift curve and a baseline. The greater the area between cumulative gain curve and baseline, the better the model is. If the cumulative gains line is closer to the top-left corner of the chart signifies greater gain and better classifier. Cumulative gain curve with respect to RDF and XGBoost algorithms are showed in Fig. 12 and 17, respectively.

*4.2.4 Lift Curve*

Lift curve [19] is an evaluation of the productiveness of a predictive model evaluated as the ratio between the results achieved with and without the predictive model. If a lift curve used, then a greater area between the lift curve and the baseline signifies the model is better. The lift curve of RDF and XGBoost algorithms are illustrated in Fig. 13 and 18, respectively.

**4.3 Prediction Model Architecture and Implementation**

The architecture of the home sale prediction model is illustrated in Fig. 9. It consists of four layers. At the top, Layer 1 includes the dataset selected 21 features and the identified target *price_high_low*. Then, Layer 2 adds the five socio-economic factors *gdp*, *cpi*, *ppi*, *hpi* and *effr*, into the prior layer of home sale dataset, increasing the features to 26 and keeping the target variable unchanged. Parallel to that, Layer 2 considers the five socio-economic factors in a separate model with the same target *price_high_low*. Both these models in Layer 2 are run in Layer 3 with the RDF and XGBoost to predict classifier features for F1 and F2. Then, in Layer 4, F1 and F2 are added into the prediction model, which now includes 28 features, and used to train the RDF and XGBoost ML model to predict the home sale price.

A brief illustration of the prediction model training is as follows. The initial accuracy of the RDF model was 84%. Further, when feature engineering was applied and included the socio-economic factors in its input the accuracy increased to 90%, but still not satisfactory. Thus, to improve the accuracy even more, mean target encoding was employed. Mean encoding is evaluated as the likelihood of the target variable, so it integrates the target feature in its encoded value. After applying mean target encoding, the accuracy of the model improved to 93.5%.

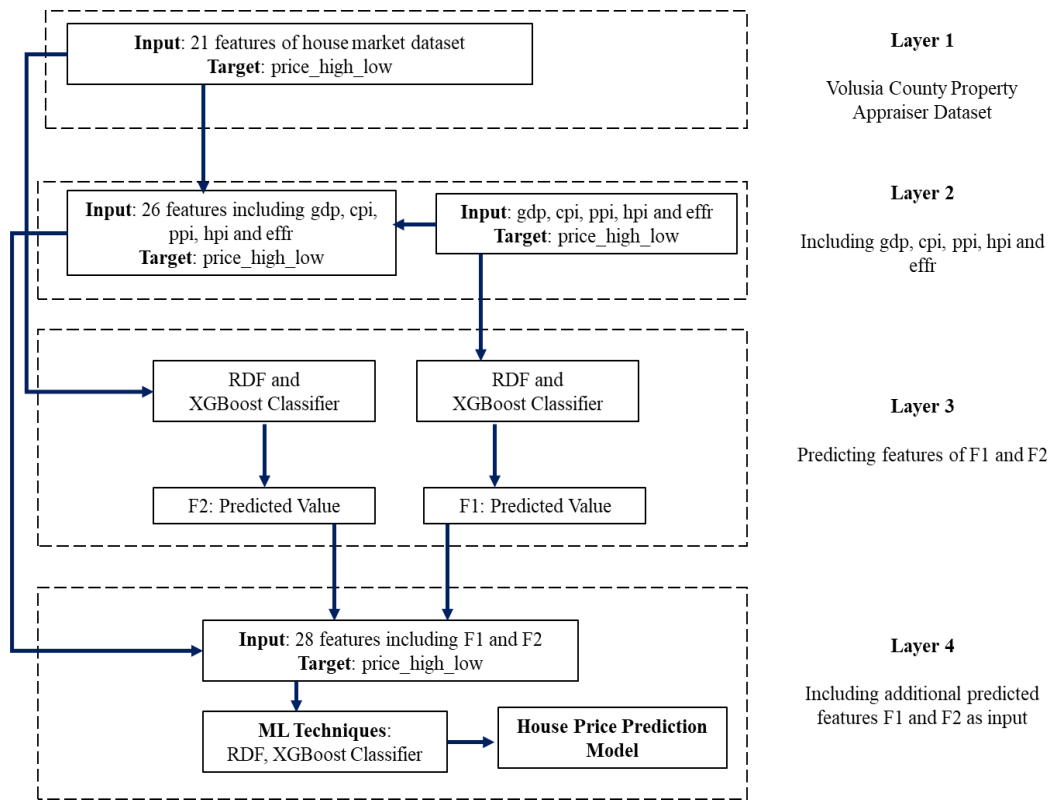

Fig. 9. Home price prediction model architecture

## 5. Model Prediction Empirical Results

The performance of running the prediction model is presented through a series of plots and discussion of the results. For the presentation of the results, classes are named as class 0 and class 1 rather than positive and negative class. Plotted charts are showing the performance of the model considering the home market features, socio-economic features and the generated features F1 and F2. Three of the performance metrics, precision, recall, and F1 score are determined via macro averaging. Both classes are considered equally important therefore the model takes the uniform average of performance metrics of both classes using the macro averaging concept.

*5.1 Random Forest Classifier*

When RDF is trained on the home sale transactions dataset with the initially identified market features, but excluding socio-economic factors, the model exhibits and accuracy of 84% and delivers almost the same value for the studied metrics, precision, recall and F1 score. By including also socio-economic factors for training purposes, the accuracy of the model significantly improves by 6% in all metrics. Finally, increasing the number of features to 28 after adding F1 and F2 and keeping all previous ones, the accuracy increases to 93.5%, and precision and recall metrics increase to 93.50%, while F1 score achieves 93%. These RDF results are also shown in Tables 2-4 along with the ones for the other three algorithms.

As shown in Fig. 10 by using the ROC curve, the value of AUC is 0.97 for both classes which indicates that RDF is well trained to make better inference for both classes. High AUC value also indicates that the classifier is better in finding the decision boundary for both the classes to make them properly separable. Moreover, K-S statistic value, shown in the plot of Fig. 12 is 0.870 at threshold value of 0.412. High K-S statistic value shows that both classes belong to different distributions with very less overlapping between both distributions. The 0.412 threshold value reveals that most of the class 0 data points fall below this threshold value, and most of the class 1 data points fall after the threshold point.

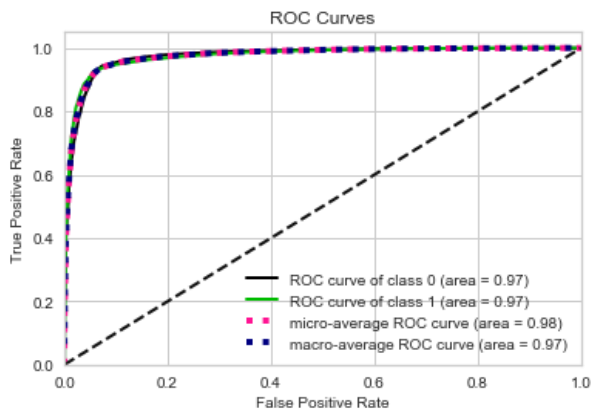
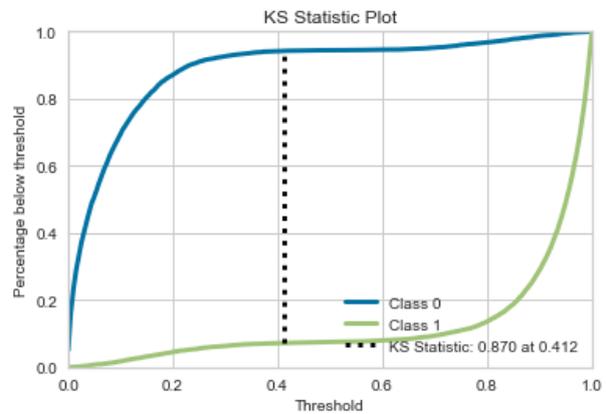

Fig. 10. ROC curve of Random Forest Classifier    Fig. 11. K-S statistic plot of Random Forest Classifier

The plot of Fig. 12 shows that the value of cumulative gain is almost the same for both classes and it reaches almost 100% at around 60% of the total sample. Then, the lift curve of Fig. 13 demonstrates that class 1 has a higher lift value than class 0. RDF can classify class 1, as making high gain compared to class 0 in earlier deciles. The lift value of 2 shows at around 40 percent of the samples, and the model makes a gain of 20% out of the total gain in all the four deciles (each decile consists 10% of population).

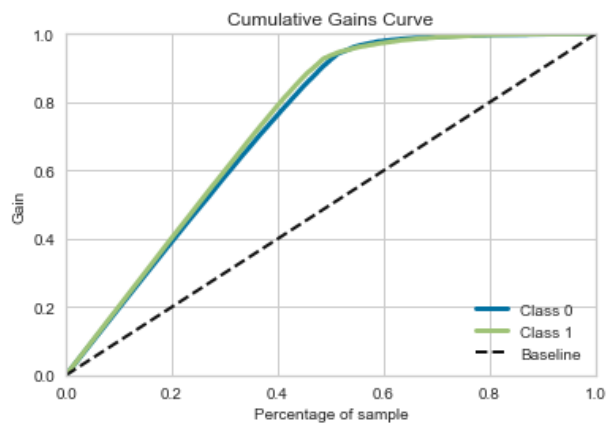
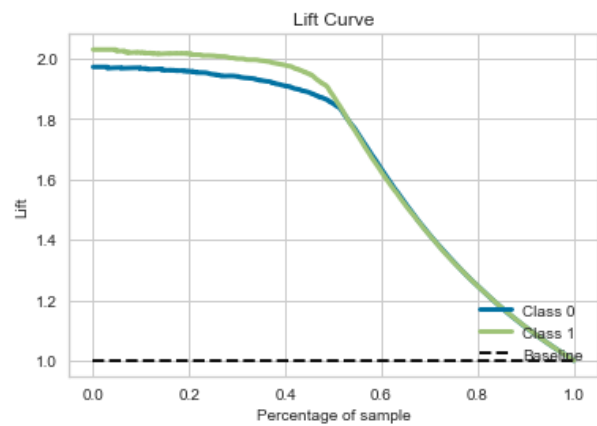

Fig. 12 Cumulative gains curve of Random Forest Classifier   Fig. 13 Lift curve of Random Forest Classifier

Using the RDF classifier report shown in Fig. 14, it can be inferred that the value of the model performance for both classes falls between 92.5-94.4%. It follows that the model is better in classifying both the classes without any bias for any class. The high F1-score of almost 93.5% for both classes provides evidence that the model is significant for both the precision and recall metrics.

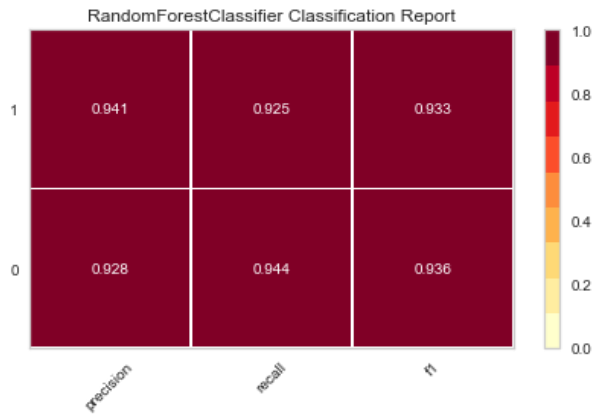

Fig. 14 Random Forest Classifier report

## 5.2. XGBoost Classifier

When XGBoost is trained on home market 21 selected features it delivers 92% accuracy with almost the same value for precision, recall and F1-score. If the identified socio-economic factors are also used for training, then a small improvement of 1% in accuracy and F1 score is obtained, but no improvement in precision and recall is observed. Finally, after adding F1 and F2 with all the previous features, the accuracy increases to 93.4%, and the value of precision, recall, and the F1 score increase to 93%.

The ROC curve depicted in Fig. 15 shows the value of AUC as 0.98 for both classes, which indicates the model is well trained to make inferences for both classes. The high AUC value also designates that the classifier is better in finding the decision boundary for both the classes and making them properly separable. The K-S statistic value of Fig. 16 is 0.873 at the threshold value of 0.305. The high K-S statistic value shows that both classes belong to different distributions with very less overlapping between both distributions, while the threshold value reveals that most of the class 0 data points fall below this threshold value, and that the class 1 data points fall after the threshold point.

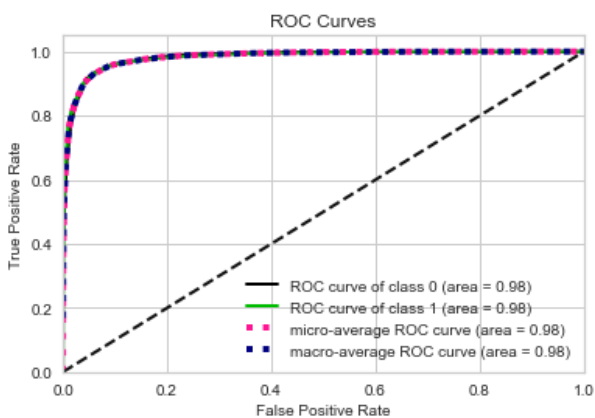 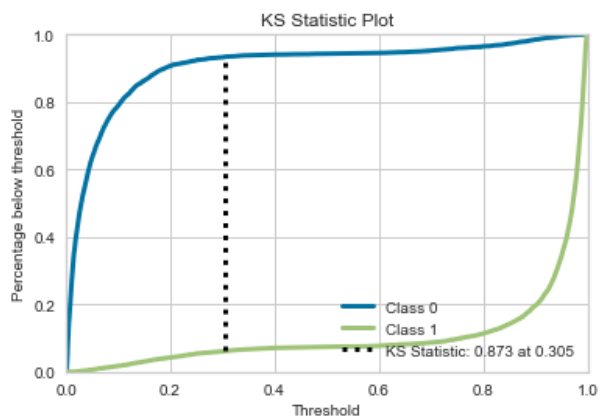

Fig. 15. ROC curve of XGBoost Classifier    Fig. 16. K-S statistic plot of XGBoost Classifier

The chart of Fig. 17 shows that the value of cumulative gain is almost the same for both classes and it reaches almost 100% at around 60% of the total sample. Then, the lift curve of Fig. 18 proves that class 1 has a higher lift value than class 0. XGBoost can classify class 1, as making high gain compared to class 0 in earlier deciles. The lift value of approximately 2 shows at up to 40 percent of the samples, and the model is able to make a gain of 20% out of the total gain in all the four deciles (each decile consists 10% of population).

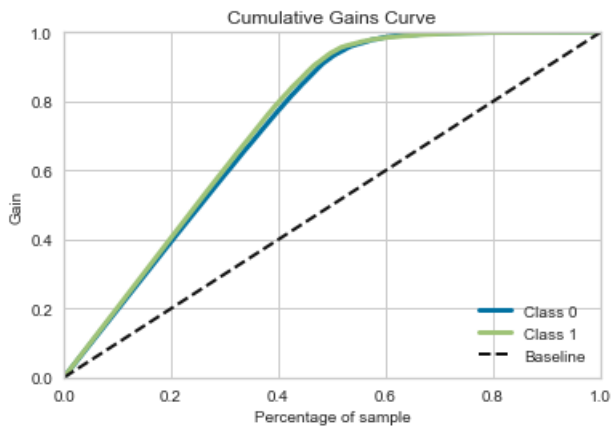
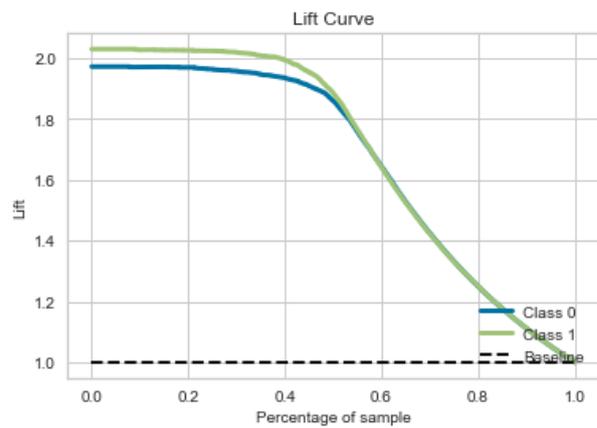

Fig. 17 Cumulative gains curve of XGBoost Classifier          Fig. 18 Lift curve of XGBoost Classifier

Using the XGBoost classifier report shown in Fig. 19, it can be inferred that the value of the model performance metric for both classes falls between 92.6-94.3%. It follows that the model is better in classifying both the classes without any bias for any class. The high F1-score of almost 93.5% for both classes provides evidence that the model is significant for both the precision and recall metrics.

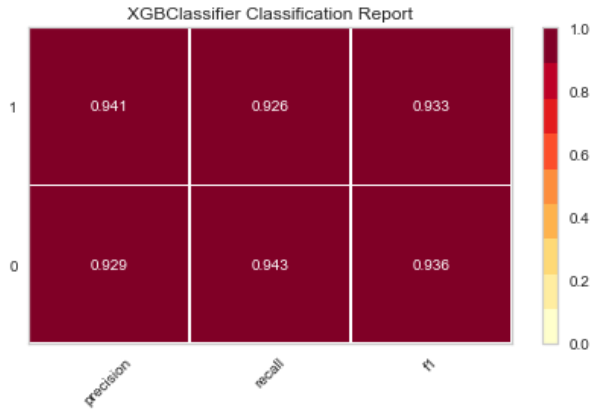

Fig. 19 XGBoost Classifier report

### 5.3. Empirical Results Analysis

The above two classifiers, RFD and XGBoost, are the two main algorithms of this study (also considered in the prediction model architecture of Section 5). Therefore, the detailed performance analysis of these two classifiers was presented in detail. The results of the remaining two classifiers, Logistic Regression and Voting Classifier are summarized in this section. All four prediction algorithms (classification models) were run following the model architecture and implementation presented in the previous section and have their results reported in Tables 2-4. To measure the performance of the classifiers, the model accuracy, precision, recall, F1 score, and error rate are determined. For a visual representation of the results, the error rate and accuracy plot for all prediction models are shown in Fig. 20-21.

| Table 2. Results without including GDP, CPI, PPI, HPI and EFFR in input Considering macro averaging (uniform metric average of both classes) | | | | | | |
|---|---|---|---|---|---|---|
| Sr. No. | Model | Accuracy | Precision | Recall | F1 | Error rate |
| 1 | Logistic | 85% | 0.85 | 0.85 | 0.85 | 15% |
| 2 | XGBoost | 92% | 0.92 | 0.92 | 0.92 | 8% |
| 3 | Random Forest | 84% | 0.84 | 0.84 | 0.84 | 16% |
| 4 | Voting | 87% | 0.87 | 0.87 | 0.87 | 13% |

The results of Table 2 are obtained by running Layer 2 prediction model with the dataset and the initially selected 21 features, without including the socio-economic factors or the mean target encoding. The best accuracy is delivered by the newer XGBoost algorithm. XGBoost also delivers the highest precision, recall, and F1 score, and it also generates the least error rate. On the lower performance side, the least accuracy, precision, recall and F1 score are given by the RDF algorithm. RDF also ranks last for the highest error rate measure. The other two classifiers, Logistic Regression and Voting provide decent results of the performance measures compared to RDF.

| Table 3. Results after including GDP, CPI, PPI, HPI and EFFR in input Considering macro averaging (uniform metric average of both classes) | | | | | | |
|---|---|---|---|---|---|---|
| Sr. No. | Model | Accuracy | Precision | Recall | F1 | Error rate |
| 1 | Logistic | 85.50% | 0.85 | 0.85 | 0.86 | 14.50% |
| 2 | XGBoost | 93% | 0.92 | 0.92 | 0.93 | 7% |
| 3 | Random Forest | 90% | 0.895 | 0.9 | 0.9 | 10% |
| 4 | Voting | 90.40% | 0.9 | 0.9 | 0.9 | 9.60% |

For the results of Table 3, the performance of the models is determined by running Layer 3 of the prediction model, which includes the five socio-economic features on top of the previously used 21 features. The highest improvements are obtained for RDF, followed by the Voting classifier. RDF accuracy increases from 84% to 90%, and the error reduces to 10%. In addition, the RDF precision, recall, and F1 score are also improved. It can also be seen that the performance of Logistic Regression and XGBoost classifiers barely improved. But still, XGBoost algorithm still significantly outperforms all other models. Major changes are observed in the results of Table 4, where two last features are added, the predicted F1 and F2. These results are obtained by running the complete prediction model, up to Layer 4 of the architecture. The Logistic Regression, Random Forest, and Voting classifier show significant improvement in performance, matching the XGBoost previous performance. The results for accuracy, precision, recall, and F1 score are almost similar irrespective of the four models.

| Table 4. Results after including predicted F1 and F2 in input Considering macro averaging (uniform metric average of both classes) | | | | | | |
|---|---|---|---|---|---|---|
| **Sr. No.** | **Model** | **Accuracy** | **Precision** | **Recall** | **F1** | **Error rate** |
| 1 | **Logistic** | 93.30% | 0.93 | 0.93 | 0.93 | 6.70% |
| 2 | **XGBoost** | 93.40% | 0.93 | 0.93 | 0.93 | 6.60% |
| 3 | **Random Forest** | 93.50% | 0.94 | 0.94 | 0.93 | 6.50% |
| 4 | **Voting** | 93.50% | 0.94 | 0.94 | 0.94 | 6.50% |

Overall, RDF and Logistic Regression algorithm show the largest improvement from one layer of the prediction model architecture to another. Also, the performance of the XGBoost model is notably better than other models in Table 2-3 and it is similar in the results shown in Table 4. Still, given the overwhelming initial better performance, the selection for the overall best performance goes to the XGBoost classifier. It also generates the least complex model, performs better in computational time, and uses lower amount of resources compared to the other three classifiers.

For visual illustration, the error rate and accuracy of each model with respect to distinct sets of features are shown in Fig 20-21. Analyzing the error rate plot of Fig. 20 for the home price prediction model, it follows that the XGBoost model produces the least error compared to other models. Then, in the accuracy plot of the prediction model shown in Fig. 21, it can be observed that the accuracy of the model in increasing order is Logistic Regression, RDF, Voting, and XGBoost.

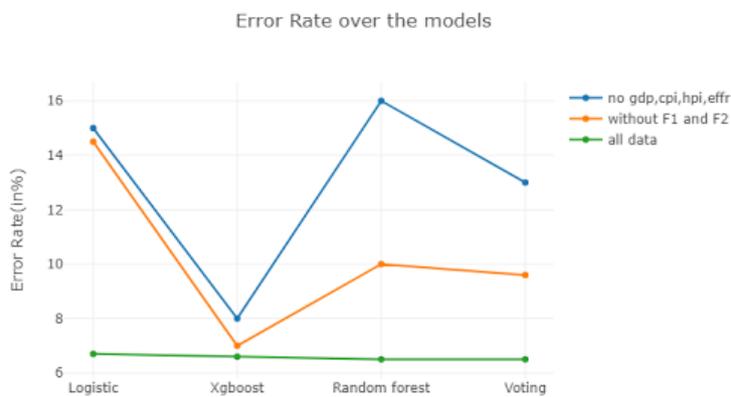

Fig. 20 Least error rate for all models

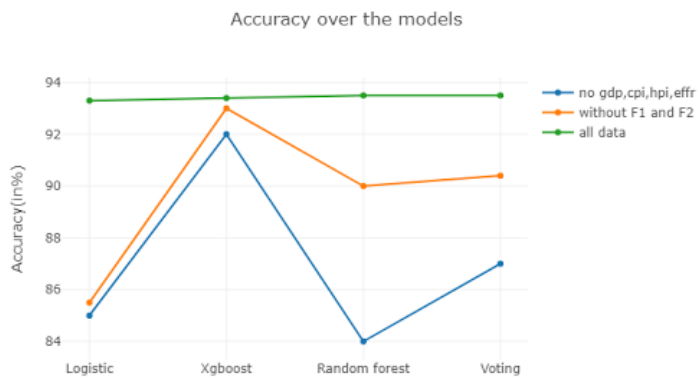

Fig. 21 Highest accuracy for all models

## 6. Conclusions and Future Work

In the real estate market, predicting the house sale price is a central problem. This study employed four ML algorithms to develop a home price prediction model to predict whether the property closing price is larger than or lower than the listing price. This is similar to evaluating whether the actual sale price of the property is greater than or lower than the appraised sale price of the property. Several types of organization can benefit from having accurate prediction models. Financial institutions can use the price prediction model to provide an accurate real estate property appraisal. Mortgage lenders can make better informed decisions for loan offers and analyze the risk factors of the real estate market. An accurate prediction model could also reduce the cost of real estate property analysis and allow faster mortgage decisions. The study also shows how the socio-economic factors of a region are highly correlated with the housing market and the importance of these factors to the real estate market.

Future work can be undertaken by the real estate market analyzers to further train the proposed algorithm or to design better ML predictors for home sale price. Datasets particularities differ from one region to another, so future studies may come up with specific solutions based on the dataset features extracted through region characteristics analysis. Nevertheless, prediction models as designed and trained work for periods of relative market calm. As experienced in 2020, worldwide events, such as viral outbreaks, can end abruptly periods of market calm. From this point of view, a clear research direction is to build ML models that can account for disruptive events. These models should be designed and trained to provide scenarios of the market outcome in the case of such disruptive events. All the stakeholders listed above would clearly benefit from market prediction models for regional or worldwide disruptive scenarios.